\documentclass[twoside]{article}
\usepackage{amssymb}
\usepackage{amsmath}
\usepackage{algorithm}
\usepackage{algpseudocode}
\usepackage{pgfplots}
\usepackage{multicol}
\usepackage[hmarginratio=1:1,top=32mm,columnsep=20pt]{geometry}
\usepackage{fullpage}
\usepackage{pdflscape}
\usepackage[toc,page]{appendix}
\usepackage[shortlabels]{enumitem}

\begin{document}
\parindent=0in
\parskip=12pt

\title{
  Sampling Bias Correction for Supervised Machine Learning \\
  \large{
    A Bayesian Inference Approach with Practical Applications
  }
}

\author{Max Sklar\\ Local Maximum Labs \\ March 11, 2022}
\date{}

\maketitle
\thispagestyle{empty}

\begin{abstract}
Given a supervised machine learning problem where the training set has been subject to a known sampling bias, how can a model be trained to fit the original dataset? We achieve this through the Bayesian inference framework by altering the posterior distribution to account for the sampling function. We then apply this solution to binary logistic regression, and discuss scenarios where a dataset might be subject to intentional sample bias such as label imbalance. This technique is widely applicable for statistical inference on big data, from the medical sciences to image recognition to marketing. Familiarity with it will give the practitioner tools to improve their inference pipeline from data collection to model selection. 
\end{abstract}

\section{Introduction}
\label{section:introduction}

Developing \textit{algorithms} is a central task in computer science. Each algorithm is a series of precise instructions to compute a given mathematical \textit{function} \(f: X \to Y\).

This works great when the programmer has these precise instructions available, but not when a function is only partially known, or if it is too complex for even a large team. \textit{Machine learning} was developed in part as a response to this problem, to extend the ultimate reach of software. 

In \textit{supervised machine learning}, an algorithm \(learns\) to compute a specific function \(f\) through example rather than through direct instructions. Examples are given in the form of \(instances\) of input-output pairs \((x, y) \in X \times Y\) known as the \textit{training set}. The training set allows the machine to learn the \textit{concept} of \(f\) by producing an approximation of it known as a \textit{model}. Models are often evaluated on a \textit{test set} of instances and finally deployed on real world data to make predictions.

Machine learning is most reliable when the training set approximates these real world conditions. Unfortunately, this is not always possible. And while models built on an imperfect dataset might perform well anyway, that performance cannot be assumed or counted upon.

When an accurate training set is unavailable either by design or circumstance, all that remains is a \textit{biased sample}. Databases that contain vast market knoweldge could have had records deleted in a non-random way. Datasets and experiments involving humans usually have sensitive information that must be redacted and filtered. Finally, vast amounts of data might be deemed unhelpful to solving the problem in question, and could be temporarily discarded in order to save costs.

\textbf{Fortunately, when the sampling function is known, there is a method that can learn the concept from the remaining evidence available, and produce a model to fit the \(original\) dataset.}

In this paper, we will fully document this method. We analyze the problem through the \textit{Bayesian framework} for supervised machine learning reviewed in section \ref{framework}. Key to this framework is the \textit{posterior probability distribution} over potential models, which allows a search algorithm to discriminate between them by their projected performance. When a biased sample is provided, the posterior distribution formula can be properly corrected as detailed in sections \ref{section:problem} and \ref{section:solution} to counteract this bias.

Unlike ad-hoc methods for correcting bias after a model has been created, this method bakes the bias-counteracting term into the learning process itself and is shown here to be consistent with first principles. It also has been successfully deployed in major commerical product launches, as documented in section \ref{section:visit}.

\section{Motivation}

While big data has unlocked incredible new applications of machine learning, the accompanying processing comes at a heavy cost in terms of time, money, and energy. Part of this cost is related to the size of the dataset being used to learn models.

This raises questions for the machine learning practitioner: Should we use all the data available to us in the training of a model? Should we take steps to procure additional data that we currently do not have? Sometimes the answer is yes. When data is scarce, each additional piece of information helps us train a better model.

But as the amount of data starts increasing by orders of magnitude, this improvement in model fit starts to diminish and the cost of training it increases. At some point the cost of obtaining and training on more data exceeds the value of the improvements to the model. This could happen very quickly on simple models (for example, a linear regression in few variables) or it could never happen (perhaps in a deep learning setup). Depending on the situation, the practitioner may choose to remove data from consideration.

One way to slim down the training set is through \textit{uniform random sampling}, where each instance in the original dataset has an equal probability of inclusion. But this process ignores the fact that some training examples are more valuable than others, particularly when there is a classification imbalance. We might therefore choose to remove data in a non-random fashion, as the following examples illustrate.

\subsection{Example: Theoretical Image Recognition}
\label{section:lion}

Suppose that the goal is to train an image recognition algorithm to detect lions. The training data contains 2,000 images with lions and 400,000 images without lions. We choose an approach to modeling these images (perhaps a convolutional neural net) and decide that we may want to sample out some of the data. Consider the following 3 sampling proposals.

\begin{enumerate}[label=(\Alph*)]
\item \label{train_on_all_data} Train on all of the data (2,000 Lions, 400,000 Non-Lions).
\item \label{train_on_randomly_sampled}Randomly sample 25 percent of the data (500 Lions, 100,000 Non-Lions).
\item \label{training_on_non_rs} Randomly sample a quarter of ONLY the non-lion images (2,000 Lions, 100,000 Non-Lions).
\end{enumerate}

Each option comes with tradeoffs.

\ref{train_on_all_data} uses all of the available data and takes the most time and resources to train. In theory, this will produce a model that is at least as good as any of the others.

\ref{train_on_randomly_sampled} has fewer instances than \ref{train_on_all_data} but with the same label imbalance. The learning algoithm will train faster, but produce worse results. A classifier relies on lion photos to learn lions, and those have become quite scarce! A reduction in lion photos from 2000 to 500 will make a significant negative impact. Because of this imbalance, each lion photo becomes much more valuable than a non-lion photo. 

\ref{training_on_non_rs}, like \ref{train_on_randomly_sampled}, demands fewer resources. The difference now is that all 2,000 lion photos remain. The likely outcome is that the model from \ref{training_on_non_rs} performs far better than \ref{train_on_randomly_sampled}. It might even perform as well as \ref{train_on_all_data}! It seems to be the best choice for balancing accuracy and efficiency.

However, because \ref{training_on_non_rs} only removed non-lion images, the derived dataset will have a \textit{bias} toward lions over non-lions. 

In the face of this bias, the practitioner is left hoping that the underlying image recognition algorithm will derive the core visual features of a lion and will not generate an excess of false positives. But a straightforward correction of the bias would certainly be preferable.

\subsection{Example: Probabilistic Event Detection}
\label{section:visit}

In the image recognition example, bias correction is desirable. In other cases, it is crucial!

An iteration of Foursquare’s attribution model\cite{visitprediction} provides a clear example. The company's attribution product measures the ability of an advertisement to drive consumers to physical locations such as a retail chain. In order to estimate causality, the Foursquare data pipeline must first learn the base probability that any given individual will visit that chain on a given day.

Foursquare's data set has many examples of visits, but examples of people who “did not visit” on any given day outnumber visits by several orders of magnitude. Therefore, it makes sense to downsample those non-visits. Because the value of an ad measurement product is in its accuracy, the sampling bias must be accounted for precisely. This was ultimately achieved through the bias correction formula for logistic regression in section \ref{section:logistic}.

\subsection{Similar and Adjacent Work}

The literature on sampling and rare events is vast. Rather than providing a comprehensive review, we refer to some selected works which were either helpful or inspirational in the course of this research.

For general cataloging of situations with imbalanced classification and techniques for its mitigation, see the work of Maalouf and Trafalis\cite{rareevents}. The readjustment formulas in the case of logistic regression can be found in Maalouf and Siddiqi\cite{weightedlogistic}. For an in-depth discussion on rare events in logistic regression, the problems associated with it, and the mathematics of parameter estimation, see King and Zeng\cite{king}.

\section{Supervised Machine Learning} \label{framework}

The following is a typical setup for supervised machine learning using a Bayesian framework, a variation of what can be found in general treatments of probabilistic inference\cite{blais}\cite{gelmanbayes}\cite{pythonbayes}. The terminology and variable names will be reused in subsequent sections.

\subsection{Given Parameters}

Let \(X\) be the input space and \(Y\) be the output space. The goal is to predict a \(label\) \(y \in Y\) from a corresponding input \(x \in X\), or to learn the function \(f: X \to Y\).

The training dataset \(D \in (X \times Y)^N\) consists of \(N\) examples of input-output pairs \((x, y) \in (X \times Y) \). \(D\) may be generated by an oracle which produces an arbitrary number of examples, or it might be a small and limited collection. In any case, inference will be based off of just these \(N\) examples.

Instances are labelled with a subscript \((x_n, y_n)\), where \(n \in \{0, 1, 2,\ldots,N-1\}\), or \(n \in N\) in ordinal notation.

\subsection{The Hypothesis Space}

Let \(H\) be the \textit{hypothesis space} whose members \(h \in H\) each encode a potential solution to the prediction problem. We assume that one of these solutions is correct. This temporary assumption makes Bayesian inference possible.

Each \(h \in H\) is considered a \textit{hypothesis} for the correct function \(f\). They are also called \textit{predictors}, \textit{models}, and \textit{solutions}. We will use the term predictor to emphasize its ultimate purpose.

In some setups, the predictors \(h \in H\) are \textit{directly predicting} \(y \in Y\) and represent a function from \(X\) to \(Y\). Here the predictors are actually \textit{probabilistically predicting} \(y \in Y\) will return a \textit{probability distribution} over \(Y\) for any given \(x \in X\). We may consider any \textit{probability measure} over \(Y\) as a valid prediction, but we keep several cases in mind.

\begin{enumerate}
	\item \(Y\) is finite. This is a \textit{classification} problem. Each predictor takes an input \(x \in X\) and returns a probability for each \(y \in Y\) that sums to one.
	\item \(Y\) is discrete but infinite. The predictor still assigns a number to each potential output, but now the infinite sum converges to one.
	\item \(Y\) is continuous. This is a \textit{regression} problem. The predictor returns a \textit{probability distribution function} (PDF) over \(Y\), which integrates to one.
\end{enumerate}

In each of these cases, the model assigns a number to each \(y \in Y\). With discrete \(Y\) it is a finite probability, and with continuous \(Y\), it is the value of the PDF. If these numbers are normalized, the discrete probabilities will add to 1 and the continous PDF will integrate to 1.

Because these values may not be normalized at first, we identify each predictor \(h\) with a \textit{relative probability function} \(f_h: X \times Y \rightarrow \mathbb{R}_{\geq 0}\) which assignes a non-negative real number to each input-output pair. Define the normalized probability function \(\hat{f}_h: X \times Y \rightarrow \mathbb{R}_{\geq 0}\) as follows for both discrete and continuous \(Y\).

\begin{align}
\label{eq:normalized_probability_model}
\hat{f}_h(x, y)=\frac{f_h(x, y)}{\sum_{y' \in Y} f_h(x,y')} &
\qquad\hat{f}_h(x, y)=\frac{f_h(x, y)}{\int_{y' \in Y} f_h(x,y')}
\end{align}

\subsection{Deriving the Prior, Likelihood, and Posterior}

The \textit{prior distribution} over \(H\) is a probability distribution representing our initial belief over which predictor is correct. Typically this will be an \textit{uninformative prior} which encodes our absense of knowledge of the problem. It is also common to incorporate \textit{Occam's Razor}, which penalizes more complex predictors, or to use a prior that is mathematically convenient. We will use \(\mathbf{P}(h)\) as the \textit{prior probability} of predictor \(h \in H\). \(\mathbf{P}(h)\) may be a discrete probability or a PDF value.

\(\mathbf{P}(h)\) does not need to be normalized. It is enough that it represents a relative probability function on \(h\) which preserves the ratio of the probabilities between two predictors. We are open to the possibility that \(\mathbf{P}(h)\) is an \textit{improper probability distribution function}. For example, a PDF that assigns 1 to each real number if \(H=\mathbb{R}\) cannot be normalized, but can still be used. More importantly, allowing the prior to be unnormalized means that we can use distributions whose integral or sum is difficult to compute.

\(\mathbf{P}(D|h)\) is the \textit{likelihood function}. It denotes the probability of recieving the entire training set \(D\) under a given predictor \(h\). We assume that the examples in the training set are \textit{independent and identically distributed} (IID). This means that the likelihood is equal to the product of the probabilities of receiving each label \(y_n\) independently.

\begin{equation}
\label{eq:likelihood_expansion}
\mathbf{P}(D|h)=\prod_{n \in N} \hat{f}_h(x_n,y_n)
\end{equation}

\(\mathbf{P}(h|D)\) is the \textit{posterior distribution} over \(H\). It represents the probability of each predictor being correct \textbf{after} the data has been taken into account.

Bayes rule for discrete \(H\) finds the posterior probability of each \(h \in H\), and in the continuous case it produces PDF values.

\begin{align}
\mathbf{P}(h|D)=\frac{\mathbf{P}(D|h)\mathbf{P}(h)}{\sum_{h' \in H}\mathbf{P}(D|h')\mathbf{P}(h')} &
\qquad\mathbf{P}(h|D)=\frac{\mathbf{P}(D|h)\mathbf{P}(h)}{\int_{h' \in H}\mathbf{P}(D|h')\mathbf{P}(h')}
\end{align}

Most learning algorithms only require an unnormalized \(\mathbf{P}(h|D)\). We rewrite the equality as a proportionality statement and remove the denominator. This form allows the prior \(\mathbf{P}(h)\) to be unnormalized as well, and covers both the discrete and continuous case.

\begin{equation}
\label{eq:bayes}
\mathbf{P}(h|D)\propto\mathbf{P}(D|h)\mathbf{P}(h)
\end{equation}

Going forward, any terms on the right hand side of this proportionality that are constant with respect to \(h\) can be removed. Using equation~\eqref{eq:likelihood_expansion}, we rewrite the formula for the relative posterior distribution as

\begin{equation}
\label{eq:bayes_likelihood_expanded}
\mathbf{P}(h|D)\propto \left( \prod_{n \in N} \hat{f}_h(x_n,y_n) \right) \mathbf{P}(h).
\end{equation}

\subsection{Selecting and Sampling Predictors}

The final learning task is to either \textit{select} or \textit{sample} predictors that best explain the data. This process involves a search of \(H\) and is identified as the search algorithm or learning algorithm.

The goal of selection is to identify a single optimal predictor. The most obvious variable to optimize is the posterior probability (or PDF value) and this is called the \textit{maximum a posteriori} (MAP) estimate. The \textit{maximum likelihood estimate} (MLE) finds the predictor that assigns the highest likelihood to the dataset, ignoring priors altogether. Selection techniques include hill climbing and gradient descent. The Newton-Raphson method is a second order method that converges faster than gradient descent.\footnote{For a satisfying example, see Fast MLE Computation for the Dirichlet Multinomial by Sklar\cite{sklar_dirichlet}.}

Samplers\footnote{This should not be confused with the data sampling, the main concern of this paper formalized in Section \ref{section:problem}.} are more complex than selectors and should be deployed selectively. The goal of the sampler is to randomly pull predictors from the posterior distribution or something approximating it. In this way we can obtain a wide variety of possible solutions that are still consistent with the data. Markov Chain Monte Carlo methods are used for sampling. A good example of this is the \textit{No U-Turn Sampler}\cite{gelman} popular in the PyMC3\cite{pymc3} probabilistic programming package for python.

These learning algorithms often use the \textit{negative log-likelihood loss} function rather than the posterior distribution directly.\footnote{For a great treatment on loss functions and their various tradeoffs, see A Tutorial on Energy Based Learning by Lecun\cite{lecun}. The work also discusses strategies for dealing with predictors where normalization over \(Y\) is not practical thus avoiding \(\hat{f_h}\).} Negative log-likelihood turns products into sums to produce values that are within a reasonable order of magnitude. We get this by applying \(-\ln(\ldots)\) to the right hand side of equation~\eqref{eq:bayes_likelihood_expanded}.

\[L(h)=-\sum_{n \in N} \ln(\hat{f}_h(x_n,y_n))-\ln(\mathbf{P}(h))\]

Let each hypothesis come with its own negative log-likelihood loss function \(l_h\) where \(\hat{f}_h(x_n,y_n)\propto e^{-l_h(x_n,y_n)}\) and the prior \(\mathbf{P}(h)\) can be reduced to a \textit{regularization function} \(\mathbf{r}(h)\) where \(\mathbf{P}(h)\propto e^{-\mathbf{r}(h)}\). The final form of the loss function is

\[L(h)=\sum_{n \in N} l_h(x_n,y_n)+\mathbf{r}(h).\]

At this point, \(L(h)\) can be used to feed various algorithms. For example, gradient descent can be used if the function has a derivative.

\pagebreak
\section{The Sampling Problem}
\label{section:problem}

What if the training set \(D\) was derived by downsampling a larger dataset \(D^+\)? Formally, we say that \(D\) was generated from \(D^+\) with a \textit{sampling probability function} \(\mathbf{s}\).

We limit ourselves to samplers \(\mathbf{s}: X \times Y \rightarrow \left [ 0, 1\right ]\) that consider each datapoint \((x, y) \in D^+\) independently.

This type of sampling is optimal for parallel computation because the sampler has no state other than it's inputs \((x, y)\). It does exclude some common sampling types, covered in Section \ref{section:future_work}.

\textbf{When the sampling function is known, a posterior distribution can still be computed from \(D\) to learn which predictors are more likely to fit \(D^+\)}.

\section{The General Solution}
\label{section:solution}

Start with the unnormalized Bayes rule in equation~\eqref{eq:bayes}, but now consider that the posterior distribution and likelihood both depend on the sampling function \(\mathbf{s}\).

\[\mathbf{P}(h|D,\mathbf{s})\propto\mathbf{P}(D|h,\mathbf{s})\mathbf{P}(h)\]

We break down the likelihood as follows.

\[\mathbf{P}(D|h,s)\propto\prod_{n \in N}\mathbf{P}(y_n|x_n,h,\mathbf{s})\]

We are left with calculating \(\mathbf{P}(y_n|x_n,h,\mathbf{s})\). The following \textit{generative description} is a useful tool in understanding how \(y_n\) is produced when there is a sampling function.

\begin{enumerate}
	\item Consider as given an input \(x_n\), a predictor \(h\), and a sampling function \(\mathbf{s}\). The predictor \(h\) encodes a probability distribution over \(Y\) through the function \(f_h(x_n,y_n)\).
	\item \label{l} Sample from that probability distribution and make this a candidate for \(y_n\), called \(y_n^*\).
	\item Compute the sampling rate \(\mathbf{s}(x_n,y_n^*)\) and use it to probabilistically determine whether \(y_n^*\) is accepted.
    \begin{enumerate}
        \item If it is accepted, return \(y_n=y_n^*\).
        \item If it is not accepted, return to step \ref{l} to generate another candidate.
    \end{enumerate}
\end{enumerate}

We now use the generative description to produce a recursive equation for \(\mathbf{P}(y_n|x_n,h,\mathbf{s})\). Let \(y \in Y\) be the first candidate for \(y_n\). If \(y\) is accepted, the probability of getting \(y_n\) is equal to 1 if \(y_n = y\) and 0 if \(y_n \neq y\). We use the indicator function notation \(\left [y_n = y\right ]\) to represent this binary output.

If \(y\) is not accepted, then the probability reverts to the original value of \(\mathbf{P}(y_n|x_n,h,\mathbf{s})\). Putting it together, the probabilty of ultimately accepting \(y_n\) comes to

\[P(y_n|cand=y,x_n,h,\mathbf{s})=\mathbf{s}(x_n,y)\left [y_n = y\right ] + (1-\mathbf{s}(x_n,y))P(y_n|x_n,h,\mathbf{s}).\]

If \(Y\) is discrete, the probability of selecting \(y\) as a candidate in the first place is \(\hat{f}_h(x_n, y)\). Sum over the probabilities of selecting each possible candidate to set up the recursive equation as follows.

\begin{equation}
\label{eq:bias_corrected_setup}
\mathbf{P}(y_n|x_n,h,\mathbf{s})=\sum_{y \in Y}\hat{f}_h(x_n,y)\big(\mathbf{s}(x_n,y)\left [y_n = y\right ] + (1-\mathbf{s}(x_n,y))P(y_n|x_n,h,\mathbf{s})\big)
\end{equation}

With the algebraic manipulation documented in appendix \ref{appendix:solving}, we solve for \(\mathbf{P}(y_n|x_n,h,\mathbf{s})\), reduce \(\hat{f}\) to \(f\), and derive the formulas for both discrete and continuous \(Y\).\footnote{The continuous version is analogous to the discrete, requiring integrating over all candidates instead of summation. Some care must be taken with the indicator function term for a rigorous argument, but ultimately it can be reduced in the same way.}

\begin{align}
\label{eq:bias_corrected_prob}
\mathbf{P}(y_n|x_n,h,\mathbf{s})=\frac{f_h(x_n,y_n)\mathbf{s}(x_n,y_n)}{\sum_{y \in Y}f_h(x_n,y)\mathbf{s}(x_n,y)} &
\qquad\mathbf{P}(y_n|x_n,h,\mathbf{s})=\frac{f_h(x_n,y_n)\mathbf{s}(x_n,y_n)}{\int_{y \in Y}f_h(x_n,y)\mathbf{s}(x_n,y)}
\end{align}

The feasibility of computing the sum or integral term depends on the structure of \(Y\), but it is at least easy when \(Y\) is finite and small enough to be enumerated by a machine.

\subsection{Formula for Negative Log-Likelihood}

Equation~\eqref{eq:bias_corrected_prob} provides all of the tools necessary to assign relative posterior probabilities to predictors. We derive a negative log likelihood-loss function starting with the relative Bayes formula.

\[\mathbf{P}(h|D,\mathbf{s})\propto\prod_{n \in N} \left[\mathbf{P}(y_n|x_n,h,\mathbf{s})\right]\mathbf{P}(h)\]

Use equation~\eqref{eq:bias_corrected_prob} to get this in terms of \(f_h\):

\[\mathbf{P}(h|D,\mathbf{s})\propto\prod_{n \in N} \left[\frac{f_h(x_n,y_n)\mathbf{s}(x_n,y_n)}{\sum_{y \in Y}f_h(x_n,y)\mathbf{s}(x_n,y)}\right]\mathbf{P}(h)\]

Finally, derive a negative log-likelihood loss function on \(h\).

\begin{equation}
\label{eq:bias_corrected_loss_function}
L(h)= \sum_{n \in N} \left[l_h(x_n,y_n)-\ln\mathbf{s}(x_n,y_n)+\ln\sum_{y \in Y}f_h(x_n,y)\mathbf{s}(x_n,y) \right] +\mathbf{r}(h)
\end{equation}

\pagebreak

\section{Solution for Binary Logistic Regression}
\label{section:logistic}

\textit{Binary logistic regression} is a special case of the supervised learning problem in section \ref{framework}. We will use it here to provide a specific example to the bias correction framework because it is particularly widespead and straightforward. Our problem is set up as follows:

\begin{enumerate}
	\item It is a \textit{binary classification} in that \(Y = \{0, 1\}\).
	\item Each input \(x \in X\) is a collection of real numerical values for each \textit{feature}. If \(F\) denotes a finite set of features, then \(X = \mathbb{R} ^{|F|}\).
	\item Each predictor \(h \in H\) is parameterized by \(h = (c, \mathbf{w})\) where \(c \in \mathbb{R}\) is the intercept and \(\mathbf{w} \in \mathbb{R}^{|F|}\) is an \(|F|\)-dimensional vector of weights corresponding to each feature.
          \item Each hypothesis \((c, \mathbf{w}) \in H\) corresponds to the following PDF, given here in both normalized and unnormalized form. In this case, \(\hat{f}_{c,\mathbf{w}}\) is a form of the \textit{sigmoid} function.
          \begin{align}
	   \label{eq:logistic_original_sigmoid}
          f_{c,\mathbf{w}}(x_n, y_n)=e^{y_n \cdot (c+\mathbf{w} \cdot x_n)} &
          \qquad\hat{f}_{c,\mathbf{w}}(x_n, y_n)=\frac{e^{y_n \cdot (c+\mathbf{w} \cdot x_n)}}{1+e^{c+\mathbf{w} \cdot x_n}}
         \end{align}
\end{enumerate}

\subsection{Formula for the Posterior Distribution}

Use equation~\eqref{eq:bias_corrected_prob} for the bias-corrected likelihood to get

\[\mathbf{P}(y_n|x_n,c,\mathbf{w},\mathbf{s})=\frac{f_{c,\mathbf{w}}(x_n,y_n)\mathbf{s}(x_n,y_n)}{\sum_{y \in Y}f_{c,\mathbf{w}}(x_n,y)\mathbf{s}(x_n,y)}=\frac{e^{y_n \cdot (c+\mathbf{w} \cdot x_n)}\mathbf{s}(x_n,y_n)}{\mathbf{s}(x_n,0)+e^{c+\mathbf{w} \cdot x_n}\mathbf{s}(x_n,1)}.\]

We introduce \(\mathbf{s}_r: X \to [0, \infty]\) as the \textit{sample ratio} for each input \(x \in X\). This ratio is all that is needed to correct for sampling bias in any binary classification. We divide through by \(\mathbf{s}(x_n,1)\) to obtain the following.

\begin{align}
\mathbf{P}(y_n|x_n,c,\mathbf{w},\mathbf{s})=\frac{e^{y_n \cdot (c+\mathbf{w} \cdot x_n)}\mathbf{s}_r(x_n)^{1 - y_n}}{\mathbf{s}_r(x_n)+e^{c+\mathbf{w} \cdot x_n}}
\qquad \textrm{where} \quad \mathbf{s}_r(x_n)=\frac{\mathbf{s}(x_n, 0)}{\mathbf{s}(x_n, 1)}
\end{align}

These problems are often framed to focus on the probability of the \textit{target condition} \(y_n = 1\). This target condition is usually the rare event that triggered the decision to use biased sampling, and would correspond to the lion image in section \ref{section:lion} and the visit in section \ref{section:visit}. We can solve for it as follows:

\begin{equation}
\label{eq:logistic_corrected_targed_prob}
\mathbf{P}(y_n = 1|x_n,c,\mathbf{w},\mathbf{s})=\frac{e^{c+\mathbf{w} \cdot x_n}}{\mathbf{s}_r(x_n)+e^{c+\mathbf{w} \cdot x_n}}
\end{equation}

Note that when \(\mathbf{s}_r(x_n) = 1\), we return to the pre-bias correction formula of equation~\eqref{eq:logistic_original_sigmoid}. This makes sense because a sampling ratio of 1 means that instances of both labels are being removed from the dataset at the same rate, thus yielding uniform random sampling.

\subsection{Finding a Suitable Loss Function}

Logistic regression is very amenable towards building a loss function. The first step is to  decide upon \(l_{c,\mathbf{w}}: X \times Y \rightarrow \mathbb{R}\) by taking the minus log of \(\mathbf{P}(y_n|x_n,c,\mathbf{w},\mathbf{s})\).

\begin{equation}
\label{eq:minus_log_prob_logistic}
-\ln \mathbf{P}(y_n|x_n,c,\mathbf{w},\mathbf{s})=\ln\left(\mathbf{s}_r(x_n)+e^{c+\mathbf{w} \cdot x_n}\right) - y_n \cdot (c+\mathbf{w} \cdot x_n) - (1 - y_n) \ln \mathbf{s}_r(x_n)
\end{equation}

Because \(\mathbf{P}(y_n|x_n,c,\mathbf{w},\mathbf{s})\) only needs to be proportional to \(e^{-l_n(x_n,y_n)}\) with respect to \(c\) and \(\mathbf{w}\), we may remove the final term in equation~\eqref{eq:minus_log_prob_logistic} to get an equivalent (and simpler) formula for \(l_{c,\mathbf{w}}\).

\begin{equation}
\label{eq:corrected_logistic_loss}
l_{c,\mathbf{w}}(x_n,y_n)=\ln\left(\mathbf{s}_r(x_n)+e^{c+\mathbf{w} \cdot x_n}\right) - y_n \cdot (c+\mathbf{w} \cdot x_n)
\end{equation}

For the regularization term, we can use \textit{L2} or \textit{ridge regression} on the weights with parameter \(\lambda\).

\begin{align}
\label{eq:l2_regularization}
\mathbf{P}(c,\mathbf{w}) \propto e^{-\frac{1}{2}\lambda(\mathbf{w} \cdot \mathbf{w})}&
\qquad\mathbf{r}(c,\mathbf{w})=\frac{1}{2}\lambda(\mathbf{w} \cdot \mathbf{w})
\end{align}

This is equivalent to putting a normal (or gaussian) prior distribution on each weight independently with a mean of \(0\) and a \textit{precision} of lambda. A large lambda produces a stronger prior which will encourage weights closer to 0. The parameter \(c\) is not given a gaussian prior here, and without it we are actually assuming a uniform, improper prior over all real numbers. It is common practice not to regularize the intercept.

Use equation~\eqref{eq:bias_corrected_loss_function} to derive the negative log-likelihood loss function by plugging in formulas from equations \eqref{eq:corrected_logistic_loss} and \eqref{eq:l2_regularization}.

\[L(c,\mathbf{w})=\sum_{n \in  N} \left (\ln\left (\mathbf{s}_r(x_n)+e^{c+\mathbf{w} \cdot x_n}\right ) -y_n \cdot (c+\mathbf{w} \cdot x_n) \right )+ \frac{1}{2}\lambda(\mathbf{w} \cdot \mathbf{w})\]

The derivative on a single weight \(\mathbf{w}_f\) or intercept \(c\) can be computed into the following simple form in order to run a gradient descent.

\[\frac{\partial}{\partial \mathbf{w}_f }L(c,\mathbf{w})=\sum_{n \in  N} x_{n,f}\left(\frac{e^{c+\mathbf{w} \cdot x_n}}{\mathbf{s}_r(x_n)+e^{c+\mathbf{w} \cdot x_n}} -y_n  \right )+ \lambda \cdot w_f\]
\[\frac{\partial}{\partial c}L(c,\mathbf{w})=\sum_{n \in  N} \left(\frac{e^{c+\mathbf{w} \cdot x_n}}{\mathbf{s}_r(x_n)+e^{c+\mathbf{w} \cdot x_n}}  -y_n\right )\]

If the target probabilities in equation~\eqref{eq:logistic_corrected_targed_prob} are pre-computed, these simplify even further!

\[\frac{\partial}{\partial \mathbf{w}_f }L(c,\mathbf{w})=\sum_{n \in  N} x_{n,f}\left(\mathbf{P}(y_n = 1|x_n,c,\mathbf{w},\mathbf{s}) -y_n  \right )+ \lambda \cdot w_f\]
\[\frac{\partial}{\partial c}L(c,\mathbf{w})=\sum_{n \in  N} \left(\mathbf{P}(y_n = 1|x_n,c,\mathbf{w},\mathbf{s})  -y_n\right )\]

\section{Future Work}
\label{section:future_work}

\subsection{Simple Random Sampling}

Our sampling function \(\mathbf{s}\) decides on the inclusion of each training instance independently. A common alternative to this is called \textit{Simple Random Sampling} (SRS) which retains a random sample with an exact number of instances. Along with this technique, the practitioner may stratify by label to correct for the imbalance of rare events. Because the sampling process is no longer independent by instance, deriving the posterior formula is now a far more complex task.

\subsection{Oversampling}

Our sampling function assumes that each datapoint is either included excluded from the dataset. This is known as \textit{undersampling}. We could allow for \textit{oversampling}, where some instances are included in \(D\) multiple times. \textit{Bootstrap sampling} methods rely on oversampling.

Now instead of \(\mathbf{s}: X \times Y \rightarrow \left [ 0, 1\right ]\), the sampling function \(\mathbf{s}\) returns a probability distribution over all natural numbers. This could be any probability distribution, but in practice it is often the \textit{poisson distribution}. The poisson distribution with parameter \(\lambda\) places probability of multiplicity \(k\) at \(\frac{\lambda^k e^{-\lambda}}{k!}\).

\subsection{The Value of an Instance}

The ability to quantify and quickly estimate the expected value that an instance will provide if included in \(D\) would allow engineers to deploy samplers that eliminate unneccesary work. Insights on sampling can be borrowed from the field of \textit{active learning}, where learners can directly choose inputs to query. Angluin\cite{angluin} provides foundational analysis on the efficiencies that can be achieved with active strategies. Many such strategies have been deployed for statistical models, and for example Cohn et al.\cite{active} propose selecting input data to minimize learner variance. Engineers must consider the specific goals and tradeoffs of a project to inform their choice of sampling strategy.

\subsection{Stochastic Gradient Descent and Mini-Batch Training}

Selection techniques like gradient descent often look at training instances one at a time or in small batches (mini-batch) while learning. This is a widespread technique that leads to enormous efficiency gains and should not be overlooked. In fact, this strategy uses the same argument as in section \ref{section:lion}. In the case of gradient descent, the exact gradient is not neccesary to take the next step, only an approximation of it. These algorithms can incorporate bias data selection and correction for additional gains.

\subsection{Ensemble Models}

Models produced from several sampling functions can be combined into an \textit{ensemble model}. They will likely perform better than in the singular approach. Because they can be run in parallel, this provides real engineering benefits.

\begin{appendices}

\section{Solving for the General Formula}
\label{appendix:solving}

Here we start with equation~\eqref{eq:bias_corrected_setup} and go through the series of steps neccesary to derive its final form in equation ~\eqref{eq:bias_corrected_prob}. Equation~\eqref{eq:bias_corrected_setup} begins as

\[\mathbf{P}(y_n|x_n,h,\mathbf{s})=\sum_{y \in Y}\hat{f}_h(x_n,y)\big(\mathbf{s}(x_n,y)\left [y_n = y\right ] + (1-\mathbf{s}(x_n,y))P(y_n|x_n,h,\mathbf{s})\big).\]

Break out the summation to get

\[\mathbf{P}(y_n|x_n,h,\mathbf{s})=\sum_{y \in Y}\hat{f}_h(x_n,y)\mathbf{s}(x_n,y)\left [y_n = y\right ] +\sum_{y \in Y}\hat{f}_h(x_n,y)(1-\mathbf{s}(x_n,y))P(y_n|x_n,h,\mathbf{s}).\]

In the first sum, the only non-zero addend is \(y = y_n\). Therefore, we can replace \(y\) with \(y_n\) and remove the summation and indicator function. In the second sum, factor out \(P(y_n|x_n,h,\mathbf{s})\) which does not contain summation index \(y\).

\[\mathbf{P}(y_n|x_n,h,\mathbf{s})=\hat{f}_h(x_n,y_n)\mathbf{s}(x_n,y_n) +P(y_n|x_n,h,\mathbf{s})\sum_{y \in Y}\hat{f}_h(x_n,y)(1-\mathbf{s}(x_n,y))\]

Collect the term \(\mathbf{P}(y_n|x_n,h,\mathbf{s})\) and distribute \(\hat{f}_h(x_n,y)\) in the remaining summation.

\[\mathbf{P}(y_n|x_n,h,\mathbf{s})\left [ 1 - \sum_{y \in Y}\hat{f}_h(x_n,y)(1-\mathbf{s}(x_n,y)) \right ]=\hat{f}_h(x_n,y_n)\mathbf{s}(x_n,y_n) \]

\[\mathbf{P}(y_n|x_n,h,\mathbf{s})\left [ 1 - \sum_{y \in Y}\hat{f}_h(x_n,y)+\sum_{y \in Y}\hat{f}_h(x_n,y)\mathbf{s}(x_n,y) \right ]=\hat{f}_h(x_n,y_n)\mathbf{s}(x_n,y_n) \]

From equation~\eqref{eq:normalized_probability_model} we know that for all possible inputs \(x\), \(\sum_{y \in Y} \hat{f}_h(x, y) = 1\). We can simplify as follows:

\[\mathbf{P}(y_n|x_n,h,\mathbf{s})\left [ 1 - 1+\sum_{y \in Y}\hat{f}_h(x_n,y)\mathbf{s}(x_n,y) \right ]=\hat{f}_h(x_n,y_n)\mathbf{s}(x_n,y_n) \]

\[\mathbf{P}(y_n|x_n,h,\mathbf{s})\left [\sum_{y \in Y}\hat{f}_h(x_n,y)\mathbf{s}(x_n,y) \right ]=\hat{f}_h(x_n,y_n)\mathbf{s}(x_n,y_n) \]

Note that \(\hat{f}_h\) is a factor on both sides of the equation, and by definition  in equation~\eqref{eq:normalized_probability_model}, \(\hat{f}_h\) is a proportional to \(f_h\). Therefore, \(\hat{f}_h\) can be reduced to \(f\), and after dividing the final form can be given as equation~\eqref{eq:bias_corrected_prob}.

\begin{equation}
\mathbf{P}(y_n|x_n,h,\mathbf{s})=\dfrac{f_h(x_n,y_n)\mathbf{s}(x_n,y_n)}{\sum_{y \in Y}f_h(x_n,y)\mathbf{s}(x_n,y)}
\tag{\ref{eq:bias_corrected_prob}}
\end{equation}

If the sampling was uniform where \(\mathbf{s}(x,y)=p\), equation~\eqref{eq:bias_corrected_prob} should reduce to the original predictor probability function \(\hat{f}_h(x_n,y_n)\) as a sanity check.

\[\mathbf{P}(y_n|x_n,h,\mathbf{s})=\frac{f_h(x_n,y_n)\mathbf{s}(x_n,y_n)}{\sum_{y \in Y}f_h(x_n,y)\mathbf{s}(x_n,y)}=\frac{f_h(x_n,y_n)p}{\sum_{y \in Y}f_h(x_n,y)p} =\hat{f}_h(x_n,y_n)\]

\end{appendices}


This document along with revisions is posted at github as https://github.com/maxsklar/bias-correction-paper. See readme for contact information. Local Maximum Labs is an ongoing effort create an disseminate knowledge on intelligent computing.

\begin{thebibliography}{20}

\bibitem{angluin}Angluin, D. (1988). Queries and concept learning. Machine learning, 2(4), 319-342.
\bibitem{blais}Blais, B. S. (2014). Statistical Inference for Everyone.
\bibitem{active}Cohn, D. A., Ghahramani, Z., \& Jordan, M. I. (1996). Active learning with statistical models. Journal of artificial intelligence research, 4, 129-145.
\bibitem{gelmanbayes}Gelman, A., Carlin, J. B., Stern, H. S., Dunson, D. B., Vehtari, A., \& Rubin, B. D. (2013). Bayesian Data Analysis. 3rd edition.
\bibitem{gelman}Hoffman, M. D., \& Gelman, A. (2014). The No-U-Turn sampler: adaptively setting path lengths in Hamiltonian Monte Carlo. J. Mach. Learn. Res., 15(1), 1593-1623.
\bibitem{king}King, G., \& Zeng, L. (2001). Logistic regression in rare events data. Political analysis, 9(2), 137-163.
\bibitem{lecun}LeCun, Y., Chopra, S., Hadsell, R., Ranzato, M., \& Huang, F. (2006). A tutorial on energy-based learning. Predicting structured data, 1(0).
\bibitem{weightedlogistic}Maalouf, M., \& Siddiqi, M. (2014). Weighted logistic regression for large-scale imbalanced and rare events data. Knowledge-Based Systems, 59, 142-148.
\bibitem{rareevents}Maalouf, M., \& Trafalis, T. B. (2011). Rare events and imbalanced datasets: an overview. International Journal of Data Mining, Modelling and Management, 3(4), 375-388.
\bibitem{pythonbayes}Martin, O. (2016). Bayesian analysis with python. Packt Publishing Ltd.
\bibitem{pymc3}Salvatier, J., Wiecki, T. V., \& Fonnesbeck, C. (2016). Probabilistic programming in Python using PyMC3. PeerJ Computer Science, 2, e55.
\bibitem{sklar_dirichlet}Sklar, M. (2014). Fast MLE computation for the Dirichlet multinomial. arXiv preprint arXiv:1405.0099.
\bibitem{visitprediction}Sklar, M., Stewart, R., Li, R., Bakula, A., \& Spears, E. (2020). Visit prediction. U.S. Patent Application No. 16/405,481. [ Section 0037 ]

\end{thebibliography}
\end{document}